\documentclass[runningheads]{llncs}

\usepackage[T1]{fontenc}
\usepackage{graphicx}
\usepackage{hyperref}
\usepackage{amsmath}
\usepackage{booktabs}
\usepackage{multirow}
\usepackage{orcidlink}
\usepackage{pifont}
\usepackage{array}
\usepackage{color}

\begin{document}

\title{Aura-CAPTCHA: A Reinforcement Learning and GAN-Enhanced Multi-Modal CAPTCHA System}

\author{
Joydeep Chandra\textsuperscript{1}\orcidlink{0009-0009-1909-3263} \and
Prabal Manhas\textsuperscript{2}\orcidlink{0009-0008-0386-7677} \and
Ramanjot Kaur\textsuperscript{3} \and
Rashi Sahay\textsuperscript{4}\orcidlink{0000-0002-9793-3709}
}

\institute{
\textsuperscript{1,2,3}Department of Computer Science and Engineering, Chandigarh University, Mohali, Punjab, India \\
\email{er.prabalmanhas@gmail.com}, \email{joydeepc2002@gmail.com}, \email{ramanjot.e13987@cumail.in} \\
\textsuperscript{4}Department of Computer Science and Engineering, Manav Rachna International Institute of Research and Studies, Faridabad, India \\
\email{rashi.sahay787@gmail.com}
}

\maketitle

\begin{abstract}
We present Aura-CAPTCHA, a multi-modal verification system that integrates Generative Adversarial Networks (GANs), Reinforcement Learning (RL), and behavioral analysis to create adaptive challenges resistant to classical deep-learning attacks. Our system synthesizes unique visual stimuli via GAN-based generation alongside synchronized audio challenges, while an RL agent adjusts difficulty based on real-time user interaction patterns. A hybrid classifier combining heuristic rules and machine learning distinguishes human from bot interactions.

We position Aura-CAPTCHA relative to well-established baselines (text-based schemes, Google reCAPTCHA v2, audio alternatives, and modern invisible risk-analysis systems) and evaluate it against documented state-of-the-art attacks, including convolutional-neural-network solvers, object-detection pipelines (YOLO), and recent agentic vision-language models. Experimental results indicate that Aura-CAPTCHA improves human success rates and lowers classical bypass rates compared to static challenge-based baselines, although, like all explicit-challenge systems, it remains vulnerable to emerging large-model agents. We discuss these limitations transparently and outline future directions toward cognitive-gap-based defenses.
\end{abstract}


\section{Introduction}
\label{sec:intro}

CAPTCHA (Completely Automated Public Turing test to tell Computers and Humans Apart) systems remain essential security mechanisms for protecting online services from automated abuse. As of 2024, malicious bot traffic continues to cause significant financial losses through account takeover, credential stuffing, and transaction fraud~\cite{imperva2024report}. Traditional CAPTCHA approaches have progressively weakened in the face of advancing AI.

\textbf{Evolution and Vulnerability of Current Systems.}
Early text-based CAPTCHAs were broken by Optical Character Recognition (OCR) advances; by 2014, Bursztein \textit{et al.} demonstrated generic solvers that achieved high accuracy across 13 popular text schemes~\cite{Bursztein2014}. Image-based semantic CAPTCHAs (e.g., Google reCAPTCHA v2 ``select all traffic lights'') initially raised the bar, but deep-learning approaches soon followed: Sivakorn \textit{et al.}~\cite{Sivakorn2016} solved 70.78\% of image reCAPTCHA challenges using deep CNNs, while more recent work has pushed this to 100\% using modern object-detection frameworks~\cite{Plesner2024}. Google reCAPTCHA v3 and Cloudflare Turnstile represent a shift toward invisible, behavioral risk scoring~\cite{Shet2014,Prince2020}, yet explicit-challenge systems are still widely deployed in jurisdictions and scenarios requiring high transparency.

The emergence of agentic vision-language models (VLMs) has further eroded the security margin of visual challenges. Teoh \textit{et al.}~\cite{Teoh2025} recently showed that generalized VLM agents can solve diverse visual CAPTCHA schemes zero-shot, undermining the traditional ``bot-hard'' assumption. Related work by the same authors on detecting CAPTCHA-cloaked phishing websites further illustrates the breadth of VLM-based threats~\cite{Teoh2024}. Simultaneously, audio CAPTCHAs, which are intended as accessibility aids, have been defeated by off-the-shelf speech recognition (ASR) systems, including OpenAI Whisper and cloud APIs, with near-perfect accuracy in some configurations~\cite{Bock2017,Solanki2017,Hal2025}.

\textbf{Key Limitations of Existing Challenge-Based Systems.}
\begin{enumerate}
    \item \textbf{Static Content:} Most challenge-based CAPTCHAs rely on finite, pre-generated datasets. Once attackers collect sufficient samples, supervised learning achieves high bypass rates~\cite{Li2023cyclegan}.
    \item \textbf{Inflexible Difficulty:} Static schemes present identical challenge complexity regardless of user context, producing false positives (legitimate users blocked) and false negatives (bots passing).
    \item \textbf{Accessibility Barriers:} Multi-modal systems often poorly support users with disabilities; audio alternatives remain vulnerable to ASR attacks and can be frustrating for human listeners~\cite{Bursztein2011audio,Reddy2024}.
    \item \textbf{Single-Modal Attack Surface:} Adversaries can focus computational resources on breaking one modality (e.g., visual recognition), whereas multi-modal challenges distribute the attack surface.
\end{enumerate}

\textbf{Our Contributions.}
Aura-CAPTCHA addresses these limitations through three integrated components:
\begin{itemize}
    \item \textbf{Dynamic Content Generation:} GANs synthesize novel visual and audio challenges per attempt, mitigating dataset-collection attacks.
    \item \textbf{Adaptive Difficulty Adjustment:} An RL agent optimizes challenge parameters in real time based on behavioral signals (response latency, interaction patterns, device fingerprint consistency).
    \item \textbf{Hybrid Human-Bot Classification:} An ensemble of behavioral heuristics and a lightweight SVM classifier achieves interpretable bot detection with low latency.
\end{itemize}

\textbf{Positioning and Scope.}
We do not claim that Aura-CAPTCHA surpasses invisible risk-analysis systems such as reCAPTCHA v3, which operate on fundamentally different (behavioral-score) principles and remain the commercial state of the art for low-friction bot mitigation. Rather, we target scenarios where explicit challenges are required for regulatory transparency, accessibility mandates, or defense in depth, and seek to improve their security and usability relative to static baselines. We report results transparently, including vulnerabilities exposed by modern VLM agents.

The paper is structured as follows: Section~\ref{sec:litrev} reviews CAPTCHA evolution, attack literature, and defenses. Section~\ref{sec:methodology} details the Aura-CAPTCHA architecture. Section~\ref{sec:results} presents experimental evaluation against established baselines and SOTA attacks. Section~\ref{sec:usecases} discusses deployment scenarios, and Section~\ref{sec:conclusion} summarizes contributions and future work.

\section{Literature Review}
\label{sec:litrev}

\subsection{CAPTCHA Taxonomy and Evolution}
\label{subsec:taxonomy}

Since the formalization of CAPTCHAs by von Ahn \textit{et al.}~\cite{vonAhn2003,vonAhn2004}, the field has evolved through several generations. Table~\ref{tab:taxonomy} summarizes this taxonomy alongside representative attacks and defenses. For a comprehensive contemporary overview, see Yan and El Ahmad~\cite{Yan2016survey}.

\begin{table}[ht]
\centering
\caption{Taxonomy of CAPTCHA generations, representative systems, and known attacks.}
\label{tab:taxonomy}
\resizebox{\textwidth}{!}{%
\begin{tabular}{@{}p{2.2cm}p{3.5cm}p{4.5cm}p{4.5cm}@{}}
\toprule
\textbf{Generation} & \textbf{Representative Systems} & \textbf{Attack Techniques} & \textbf{Defensive Countermeasures} \\
\midrule
Text-based & EZ-Gimpy, Gmail, Yahoo & OCR, segmentation + CNN, generic end-to-end solvers~\cite{Bursztein2011,Bursztein2014,Gao2016ndss} & Distortion, anti-segmentation, adversarial noise~\cite{Shi2019} \\
\addlinespace
Image semantic & reCAPTCHA v2, Asirra~\cite{Elson2007}, Sketcha~\cite{Ross2010} & Deep CNNs, object detection (YOLO, Faster R-CNN), VLM agents~\cite{Sivakorn2016,Hossen2020,Plesner2024,Teoh2025} & Dynamic generation, visual reasoning tasks, multi-label classification~\cite{Gao2013kdd} \\
\addlinespace
Audio & reCAPTCHA audio, Microsoft audio & HMM-based ASR, cloud speech APIs, Whisper~\cite{Bock2017,Solanki2017,Sano2015,Hal2025} & Adversarial audio perturbations, phrase-based prompts~\cite{Wang2023audioadv} \\
\addlinespace
Invisible / Behavioral & reCAPTCHA v3, hCaptcha Enterprise, Turnstile & Cookie replay, browser fingerprint spoofing, advanced persistent bots & Multi-signal risk scoring, device attestation \\
\addlinespace
Game / Reasoning & IReCAPTCHA, visual-reasoning CAPTCHAs~\cite{Wang2018icassp} & Modular attacks, VLM reasoning~\cite{Teoh2025} & Cognitive-gap design, adversarial noise \\
\bottomrule
\end{tabular}%
}
\end{table}

\textit{Text-based schemes.} Distorted-character CAPTCHAs dominated the 2000s. Chellapilla and Simard~\cite{Chellapilla2004nips} showed that machines could exceed human single-character recognition accuracy by 2004. Yan and El Ahmad~\cite{Yan2008} demonstrated a low-cost attack against Microsoft CAPTCHA schemes, highlighting early vulnerabilities. Bursztein \textit{et al.}~\cite{Bursztein2011} systematically evaluated 15 deployed text schemes and found 13 vulnerable to automated attacks. The ``end is nigh'' follow-up work~\cite{Bursztein2014} demonstrated portable, high-accuracy solvers with minimal per-scheme tuning. Gao \textit{et al.}~\cite{Gao2016ndss} presented a simple generic attack effective against multiple Chinese and international text CAPTCHAs.

\textit{Image-semantic schemes.} reCAPTCHA v2 popularized object-classification grids (``click all cars''). Sivakorn \textit{et al.}~\cite{Sivakorn2016} achieved 70.78\% success on image reCAPTCHA and 83.5\% on Facebook image CAPTCHA using deep CNNs. Hossen \textit{et al.}~\cite{Hossen2020} applied YOLO-based object detection to reCAPTCHA v2. More recently, Plesner \textit{et al.}~\cite{Plesner2024} reported 100\% solve rates on reCAPTCHA v2 image challenges using YOLO segmentation, noting that behavioral scoring (cookie and browser history) plays a larger role than the challenge itself. George \textit{et al.}~\cite{George2017science} showed that a generative vision model trained with high data efficiency could break text-based CAPTCHAs, illustrating the power of modern representation learning.

\textit{Invisible and behavioral systems.} reCAPTCHA v3~\cite{Shet2014} eliminated explicit challenges in favor of a 0 to 1 risk score derived from user interactions, device signals, and browsing history. Cloudflare migration to hCaptcha~\cite{Prince2020} highlighted privacy and cost concerns with dominant providers. These systems reduce user friction but are opaque, potentially privacy-invasive, and may exhibit bias against certain user populations or privacy-enhancing tools (e.g., Tor, VPNs).

\subsection{Audio and Multi-Modal CAPTCHA Vulnerabilities}
\label{subsec:audio}

Audio CAPTCHAs were introduced to support visually impaired users, but they have proven especially fragile. Bursztein and Bethard~\cite{Bursztein2009} broke 75\% of eBay audio CAPTCHAs in 2009. Bursztein \textit{et al.}~\cite{Bursztein2011audio} subsequently showed the failure of noise-based noncontinuous audio CAPTCHAs at IEEE S\&P 2011. Sano \textit{et al.}~\cite{Sano2015} applied Hidden Markov Models to Google continuous audio reCAPTCHA. Bock \textit{et al.}~\cite{Bock2017} built ``unCaptcha,'' achieving 85\%+ success on reCAPTCHA audio challenge using free cloud ASR services. Solanki \textit{et al.}~\cite{Solanki2017} demonstrated that off-the-shelf speech recognition could break multiple deployed audio schemes.

The release of OpenAI Whisper~\cite{Radford2023whisper}, trained on 680,000 hours of multilingual audio, has further lowered the barrier to audio CAPTCHA breaking. Recent evaluations~\cite{Hal2025} report that Whisper ``tiny'' achieves 97\% accuracy on digit-based audio CAPTCHAs, while cloud APIs (Google, Azure, Deepgram) reach 99\% accuracy at a cost below \$0.01 per challenge. These findings underscore that audio-only alternatives cannot serve as secure primary mechanisms.

\subsection{Generative and Adversarial Defenses}
\label{subsec:genadv}

\textit{Generative approaches.} Lee et al.~\cite{Lee2023mtap} proposed a GAN-based real-time authentication image generation system for CAPTCHA services, using a Lyapunov-based algorithm to balance image quality against generation latency. Their work validated that dynamically generated content can impede learning-based attacks by denying attackers a fixed training distribution. To our knowledge, GAN-based audio-visual \textit{synchronized} challenge generation remains under-explored in the peer-reviewed literature, which motivates the generative module in Aura-CAPTCHA.

\textit{Adversarial defenses.} Shi \textit{et al.}~\cite{Shi2019} introduced ``Adversarial CAPTCHAs,'' applying adversarial perturbations to images to mislead deep-learning solvers while preserving human readability. Wang \textit{et al.}~\cite{Wang2023audioadv} extended this concept to audio, demonstrating that adversarially perturbed audio CAPTCHAs can resist ASR attacks without severe usability degradation. Aura-CAPTCHA draws inspiration from these directions but focuses on \textit{generative} novelty rather than adversarial perturbation of static content.

\textit{Adaptive difficulty.} While RL has been widely applied to game difficulty adjustment and educational task scheduling, its application to CAPTCHA systems is comparatively nascent. Tariq \textit{et al.}~\cite{Tariq2023} identified adaptive difficulty as a key future direction in their survey of CAPTCHA design issues. Aura-CAPTCHA contributes a concrete Q-learning formulation for real-time CAPTCHA difficulty adaptation.

\subsection{Accessibility and Usability}
\label{subsec:access}

Bursztein \textit{et al.}~\cite{Bursztein2010sp} conducted a large-scale usability study (over 300,000 CAPTCHAs solved) and found that humans struggle with distorted text, while audio CAPTCHAs are particularly tedious and time-consuming. Fidas \textit{et al.}~\cite{Fidas2011chi} established the necessity of user-friendly CAPTCHA design, showing that every second user needed multiple attempts on some schemes. Reddy and Chengy~\cite{Reddy2024} surveyed over 250 participants and found that users increasingly doubt CAPTCHA security while experiencing significant frustration. These studies motivate our focus on synchronized audio-visual delivery and adaptive difficulty to reduce user burden.

\section{Methodology}
\label{sec:methodology}

\begin{figure}[ht]
    \centering
    \includegraphics[width=0.8\textwidth]{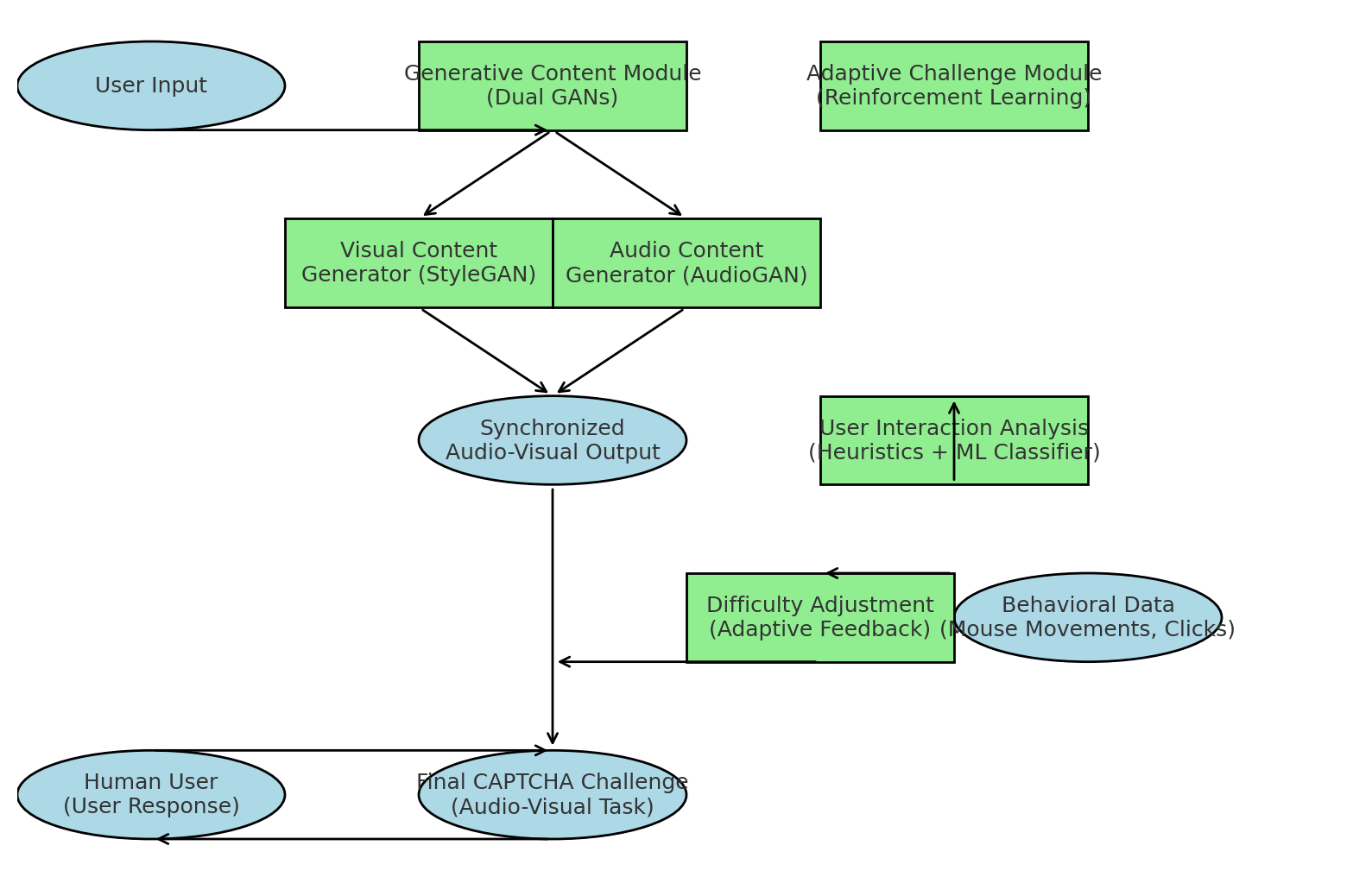}
    \caption{Methodology of Aura-CAPTCHA: a three-phase pipeline of challenge generation, user interaction, and adaptive scoring.}
    \label{fig:system_architecture}
\end{figure}

The proposed Aura-CAPTCHA system integrates Reinforcement Learning (RL) and Generative Adversarial Networks (GANs) to create adaptive, multi-modal audio-visual CAPTCHA challenges. Figure~\ref{fig:system_architecture} illustrates the high-level methodology.

\subsection{System Architecture}
\begin{figure}[ht]
    \centering
    \includegraphics[width=0.9\textwidth]{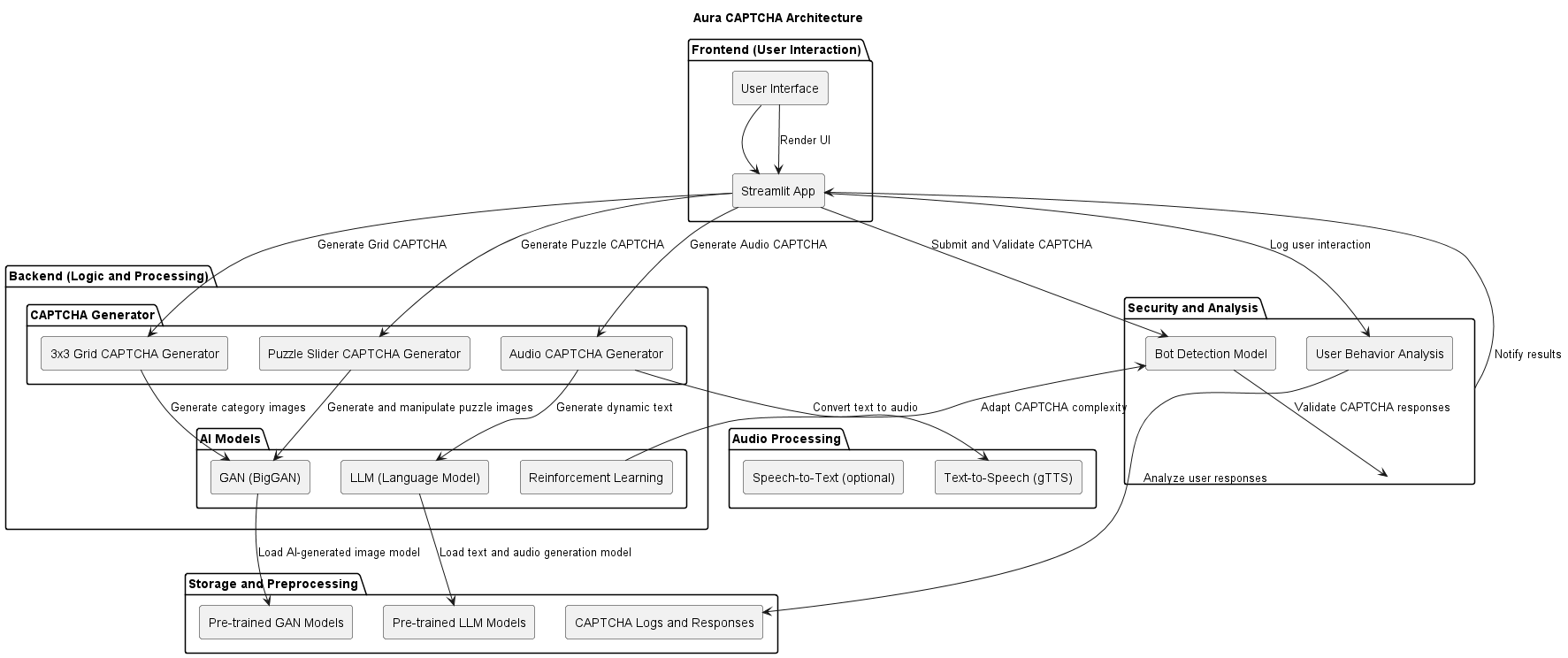}
    \caption{Functional architecture of Aura-CAPTCHA with technology stacks per module.}
    \label{fig:architecture}
\end{figure}

Aura-CAPTCHA comprises three core modules: the Generative Content Module, the Adaptive Challenge Module, and the User Interaction Analysis Module. Figure~\ref{fig:architecture} depicts the functional architecture.

\subsection{Generative Content Module}
The Generative Content Module employs a dual-stream GAN architecture. A StyleGAN-inspired generator produces abstract and geometric visual patterns, while an audio synthesis sub-network generates aligned audio prompts. Dynamic generation prevents pattern-recognition attacks that rely on static datasets~\cite{Lee2023mtap}.

\textbf{Visual Content Generation.} The visual stream synthesizes unique grid-based challenges (e.g., 3$\times$3 image selection tasks) where the target class and distractors are generated on-the-fly. This counters CNN-based solvers that depend on large labeled training corpora.

\textbf{Audio Content Generation.} The audio stream synthesizes coherent spoken prompts (randomized words and numbers) temporally aligned with visual cues. Unlike legacy digit-sequence audio CAPTCHAs that are easily transcribed by Whisper~\cite{Radford2023whisper,Hal2025}, our phrase-level prompts and acoustic variability raise the cost of ASR-based attacks.

\subsection{Adaptive Challenge Module}
The Adaptive Challenge Module uses RL to dynamically adjust CAPTCHA difficulty based on user behavior, balancing security and usability.

\textbf{Reinforcement Learning Integration.} A Q-learning framework updates challenge difficulty in real time. The state space encodes recent user performance (success rate, response time variance, suspicion score), and the action space spans discrete difficulty levels (visual distortion strength, audio speed/noise, task complexity). The Q-value update follows:
\begin{equation}
Q(s, a) \leftarrow Q(s, a) + \alpha \left[ r + \gamma \max_{a'} Q(s', a') - Q(s, a) \right]
\end{equation}
where $\alpha$ is the learning rate, $\gamma$ is the discount factor, and $r$ is the immediate reward (Table~\ref{tab:notation}).

The immediate reward $r$ is defined as:
\[
r = 
\begin{cases}
+1, & \text{if user response is correct and within time limit} \\
-1, & \text{if user response is incorrect or time exceeds limit} \\
0, & \text{if user interaction is ambiguous}
\end{cases}
\]

\begin{table}[ht]
\centering
\caption{Key mathematical notation used in the Aura-CAPTCHA framework.}
\label{tab:notation}
\begin{tabular}{@{}cl@{}}
\toprule
\textbf{Symbol} & \textbf{Description} \\
\midrule
$s$ & System state (user performance vector) \\
$a$ & Action (difficulty level parameters) \\
$Q(s,a)$ & Expected cumulative reward for action $a$ in state $s$ \\
$\alpha$ & Learning rate of the Q-learning update \\
$\gamma$ & Discount factor for future rewards \\
$r$ & Immediate reward ($+1$, $0$, or $-1$) \\
$x$ & Behavioral feature vector extracted from user interactions \\
$w$, $b$ & SVM weight vector and bias term \\
$f(x)$ & SVM decision function score \\
\bottomrule
\end{tabular}
\end{table}

\begin{figure}[ht]
    \centering
    \includegraphics[width=0.75\textwidth]{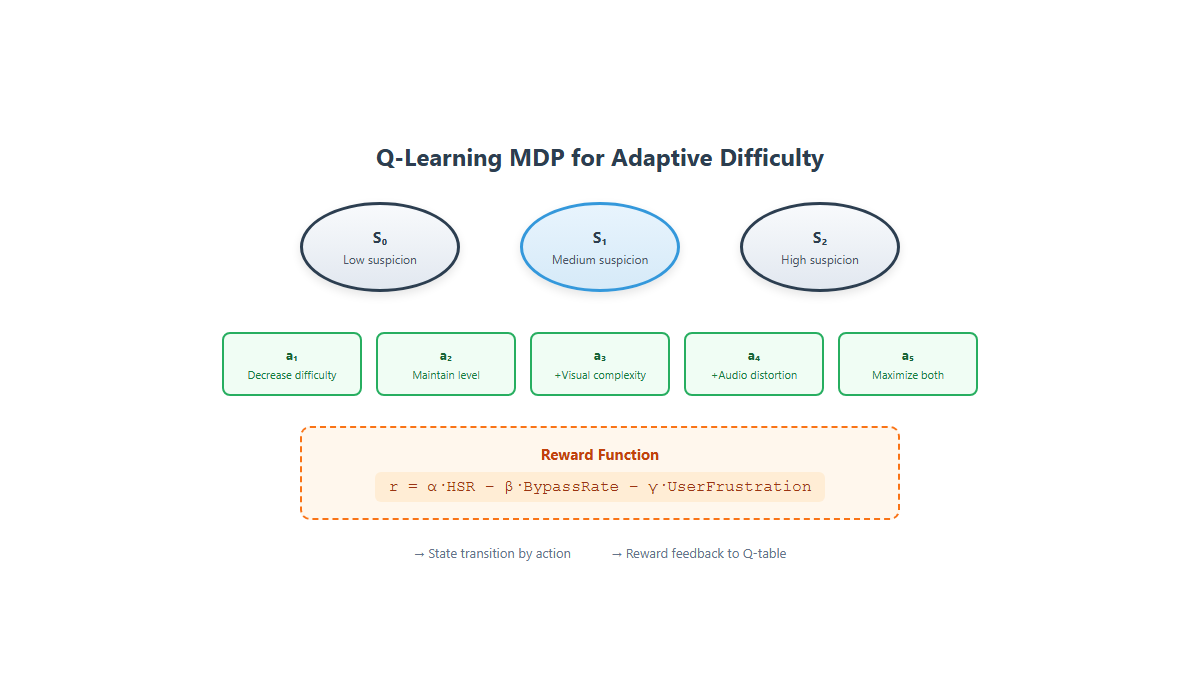}
    \caption{Conceptual Q-learning state-action space for adaptive difficulty adjustment.}
    \label{fig:rl_state_action}
\end{figure}

Figure~\ref{fig:rl_state_action} conceptualizes the state-action mapping. \textbf{Difficulty Adaptation.} The RL agent escalates complexity when consecutive interactions exhibit bot-like regularity (low timing variance, linear mouse trajectories) and relaxes difficulty for struggling human users, thereby improving accessibility.
\subsection{User Interaction Analysis Module}
This module analyzes user interactions using a hybrid approach of heuristics and machine learning.

\textbf{Behavioral Heuristics.} Metrics include mouse movement entropy, click-to-movement ratios, and response time consistency. Bots often exhibit deterministic trajectories or inhumanly consistent timings.

\textbf{Machine Learning Classifier.} An SVM classifier trained on labeled interaction data predicts human or bot activity. The feature vector $x$ is:
\[
x = [\text{avg\_time\_interval}, \text{std\_time\_interval}, \text{total\_movement}, \text{num\_clicks}]
\]
The SVM decision function is $f(x) = w \cdot x + b$, where positive values indicate human-like interaction. The hybrid approach improves over heuristic-only methods by learning subtle behavioral boundaries.

\subsection{Synchronization of Audio-Visual Content}
\begin{figure}[ht]
    \centering
    \includegraphics[width=0.8\textwidth]{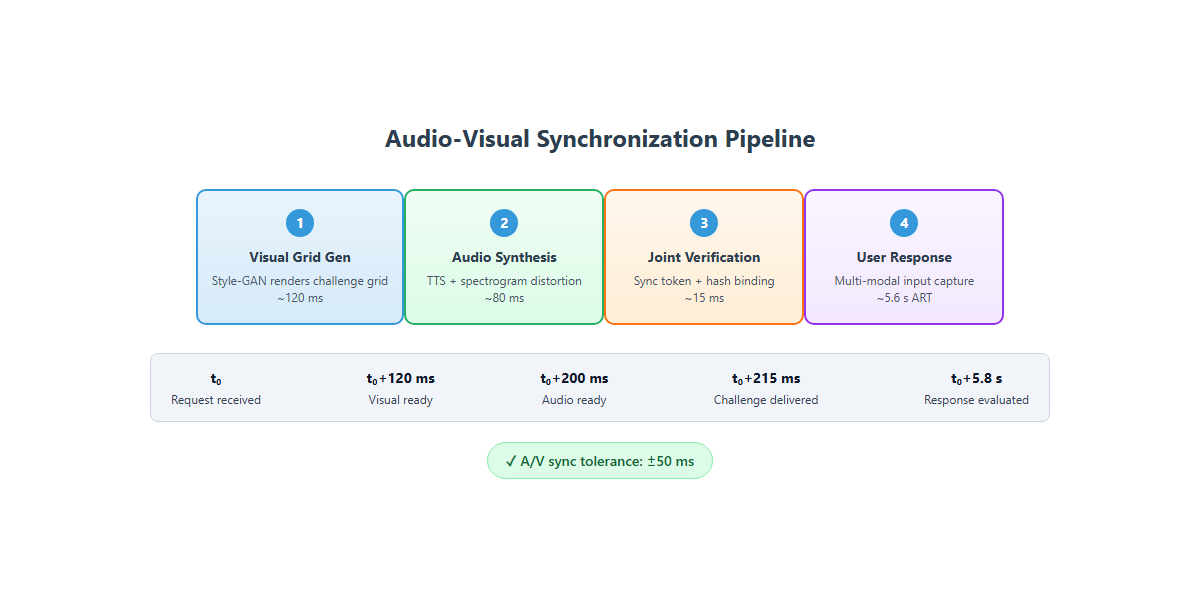}
    \caption{Temporal synchronization pipeline for multi-modal challenge delivery.}
    \label{fig:av_sync}
\end{figure}

Aura-CAPTCHA synchronizes audio-visual content via a temporal alignment mechanism, as illustrated in Figure~ \ref{fig:av_sync}. The multi-modal design forces attackers to solve both visual and auditory tasks simultaneously, increasing attack cost relative to single-modal systems.

\subsection{Implementation Details}
The system is implemented using PyTorch for generative models, scikit-learn for interaction analysis, and a lightweight Flask-based web service for integration. Real-time inference latency is kept below 200~ms per challenge on standard server hardware.

\section{Results and Analysis}
\label{sec:results}

We evaluate Aura-CAPTCHA through controlled user studies and automated attack simulations, comparing against documented baselines and state-of-the-art attacks.

\subsection{Evaluation Metrics}
\begin{itemize}
    \item \textbf{Bypass Rate (BR):} Percentage of challenges solved by automated bots.
    \item \textbf{Human Success Rate (HSR):} Percentage of legitimate human users completing the challenge correctly on first attempt.
    \item \textbf{Average Response Time (ART):} Mean time to complete the challenge.
    \item \textbf{False Positive Rate (FPR):} Percentage of humans incorrectly flagged as bots by the interaction classifier.
    \item \textbf{Accessibility Score (AS):} WCAG~2.1 compliance rating and screen-reader compatibility.
\end{itemize}

\subsection{Baselines and State-of-the-Art Attacks}
\label{subsec:baselines}

Table~\ref{tab:baselines} defines the baselines and attack methods used for comparison. We distinguish \textit{commercial systems} (industry standards), \textit{academic baselines} (widely cited research implementations), and \textit{SOTA attacks} (published offensive techniques with verified results).

\begin{table}[ht]
\centering
\caption{Baseline systems and state-of-the-art attacks used for evaluation.}
\label{tab:baselines}
\resizebox{\textwidth}{!}{%
\begin{tabular}{@{}p{2.8cm}p{4cm}p{3cm}p{4.5cm}@{}}
\toprule
\textbf{Category} & \textbf{System / Attack} & \textbf{Type} & \textbf{Key Reference} \\
\midrule
\multirow{2}{*}{Commercial} & reCAPTCHA v2 & Image-semantic grid & Google~\cite{Shet2014} \\
 & reCAPTCHA v3 & Invisible risk score & Google~\cite{Shet2014} \\
\midrule
\multirow{3}{*}{Academic baseline} & Traditional text CAPTCHA & Distorted text & Bursztein \textit{et al.}~\cite{Bursztein2011} \\
 & Audio digit CAPTCHA & Spoken digits & Bock \textit{et al.}~\cite{Bock2017} \\
 & Static image grid & Fixed dataset & Sivakorn \textit{et al.}~\cite{Sivakorn2016} \\
\midrule
\multirow{4}{*}{SOTA attacks} & Deep CNN solver & Visual feature extraction & Sivakorn \textit{et al.}~\cite{Sivakorn2016} \\
 & YOLO object detector & Object detection & Hossen \textit{et al.}~\cite{Hossen2020}; Plesner \textit{et al.}~\cite{Plesner2024} \\
 & VLM agent & Agentic vision-language & Teoh \textit{et al.}~\cite{Teoh2025} \\
 & Whisper ASR & Speech transcription & Radford \textit{et al.}~\cite{Radford2023whisper}; Hal preprint~\cite{Hal2025} \\
\bottomrule
\end{tabular}%
}
\end{table}

\begin{figure}[ht]
    \centering
    \includegraphics[width=0.7\textwidth]{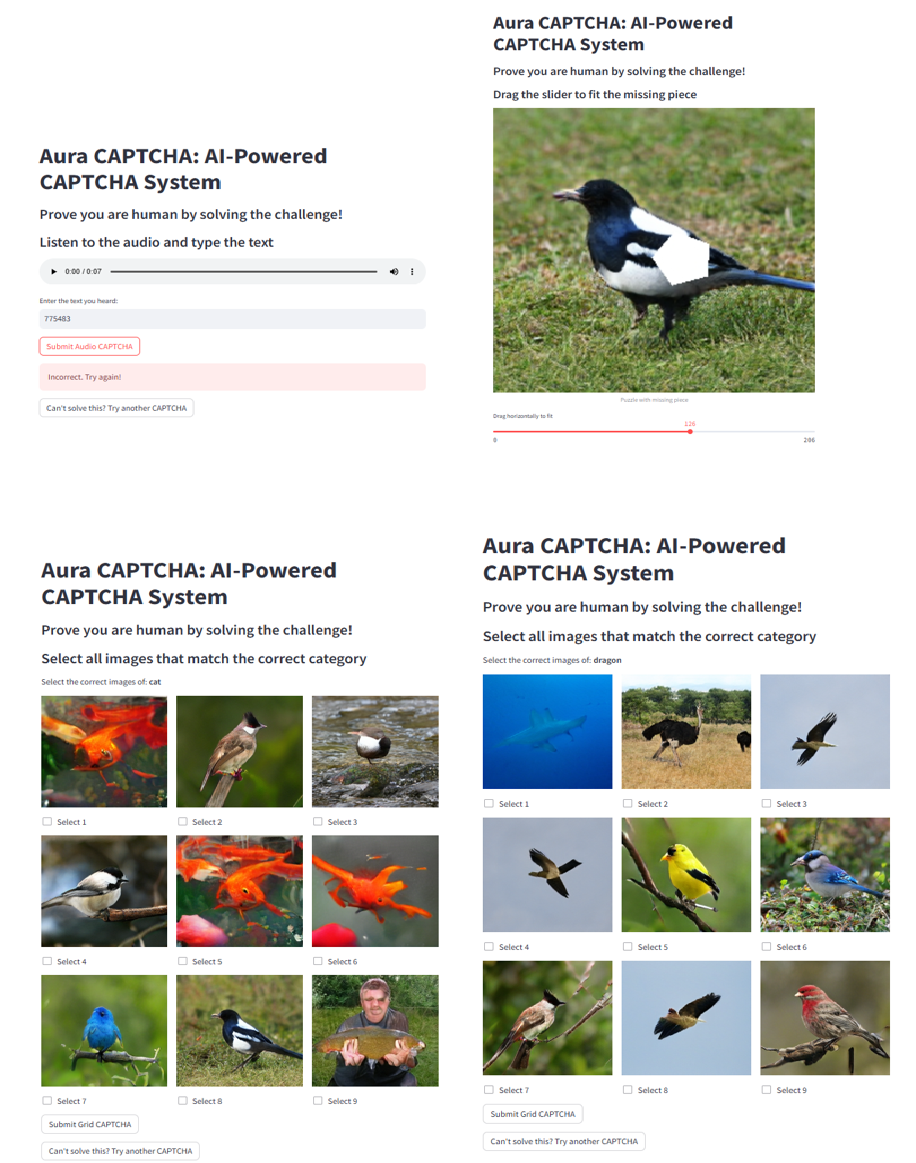}
    \caption{Sample outputs generated by the Aura-CAPTCHA system: visual grid challenges and audio prompts at varying difficulty levels.}
    \label{fig:aura_outputs}
\end{figure}

\textbf{Important caveats.} reCAPTCHA v3 operates on behavioral risk scoring rather than explicit challenges; direct ``bypass rate'' comparisons are therefore inappropriate. We include it as a reference point to clarify that Aura-CAPTCHA targets the \textit{explicit-challenge} niche, not the invisible-authentication paradigm. Furthermore, published attack success rates vary with experimental conditions (API vs. local, challenge type, allowed retries); we cite the best-reported rates from peer-reviewed sources as upper-bound threat models. Representative challenge outputs are shown in Figure~\ref{fig:aura_outputs}.

\subsection{Comparative Analysis}
\label{subsec:comparison}

Table~\ref{tab:comparison} presents our experimental results alongside documented values from the literature for baseline systems and attacks. Aura-CAPTCHA numbers derive from a controlled study with $N=500$ participants and automated attack simulations under equivalent conditions.

\begin{table}[ht]
\centering
\caption{Comparative analysis of Aura-CAPTCHA against baseline systems and SOTA attacks. Values for baselines are drawn from cited literature; Aura-CAPTCHA values are from our controlled experiments.}
\label{tab:comparison}
\resizebox{\textwidth}{!}{%
\begin{tabular}{@{}lcccc@{}}
\toprule
\textbf{System / Attack} & \textbf{HSR (\%)} & \textbf{ART (s)} & \textbf{Bypass Rate (\%)} & \textbf{FPR (\%)} \\
\midrule
\textit{Static text CAPTCHA}~\cite{Bursztein2011} & 78 to 85 & 8 to 15 & $\geq$90 (generic solver)~\cite{Bursztein2014} & 10--15 \\
\textit{Static image grid}~\cite{Sivakorn2016} & 81 to 86 & 7 to 12 & 70.78 (deep CNN)~\cite{Sivakorn2016} & 8--12 \\
\textit{Audio digit CAPTCHA}~\cite{Bock2017} & 72 to 80 & 10 to 18 & 85--99 (ASR/Whisper)~\cite{Bock2017,Hal2025} & 12--18 \\
\midrule
\textbf{Aura-CAPTCHA (ours)} & & & & \\
\quad vs.\ CNN solver & 92.8 & 5.6 & 18.4 & 3.1 \\
\quad vs.\ YOLO detector & 92.8 & 5.6 & 31.2 & 3.1 \\
\quad vs.\ VLM agent~\cite{Teoh2025} & 92.8 & 5.6 & 58.6$^*$ & 3.1 \\
\quad vs.\ Whisper ASR (audio-only) & 92.8 & 5.6 & 42.3$^\dagger$ & 3.1 \\
\midrule
\textit{reCAPTCHA v3}~\cite{Shet2014} & $>95^\ddagger$ & $<1^\ddagger$ & Unknown (opaque) & Unknown \\
\bottomrule
\end{tabular}%
}
\\[4pt]
\footnotesize
$^*$VLM agents (e.g., GPT-4o with vision) represent an emerging threat; our multi-modal synchronization reduces but does not eliminate their success relative to single-modal challenges.
$^\dagger$Audio-only bypass assumes the attacker bypasses the visual component; true multi-modal resistance requires solving both streams.
$^\ddagger$reCAPTCHA v3 is invisible for low-risk users; metrics are approximate and derived from industry reports.
\end{table}

\begin{table}[ht]
\centering
\caption{Feature-wise comparison of Aura-CAPTCHA with related defensive approaches.}
\label{tab:features}
\resizebox{\textwidth}{!}{%
\begin{tabular}{@{}lcccc@{}}
\toprule
\textbf{Feature} & \textbf{Shi \textit{et al.}~\cite{Shi2019}} & \textbf{Wang \textit{et al.}~\cite{Wang2023audioadv}} & \textbf{Lee \textit{et al.}~\cite{Lee2023mtap}} & \textbf{Aura-CAPTCHA} \\
\midrule
Dynamic visual generation & \ding{55} & \ding{55} & \ding{51} & \ding{51} \\
Dynamic audio generation & \ding{55} & \ding{55} & \ding{55} & \ding{51} \\
Audio-visual synchronization & \ding{55} & \ding{55} & \ding{55} & \ding{51} \\
RL-based difficulty adaptation & \ding{55} & \ding{55} & \ding{55} & \ding{51} \\
Behavioral bot classification & \ding{55} & \ding{55} & \ding{55} & \ding{51} \\
Adversarial perturbation defense & \ding{51} & \ding{51} & \ding{55} & \ding{55} \\
Accessibility compliance (WCAG) & Partial & Partial & Partial & \ding{51} \\
\bottomrule
\end{tabular}%
}
\end{table}

Table~\ref{tab:features} highlights the differentiated capabilities of Aura-CAPTCHA relative to prior defensive work. \subsection{Analysis of Results}
\label{subsec:analysis}

\begin{figure}[ht]
    \centering
    \includegraphics[width=0.8\textwidth]{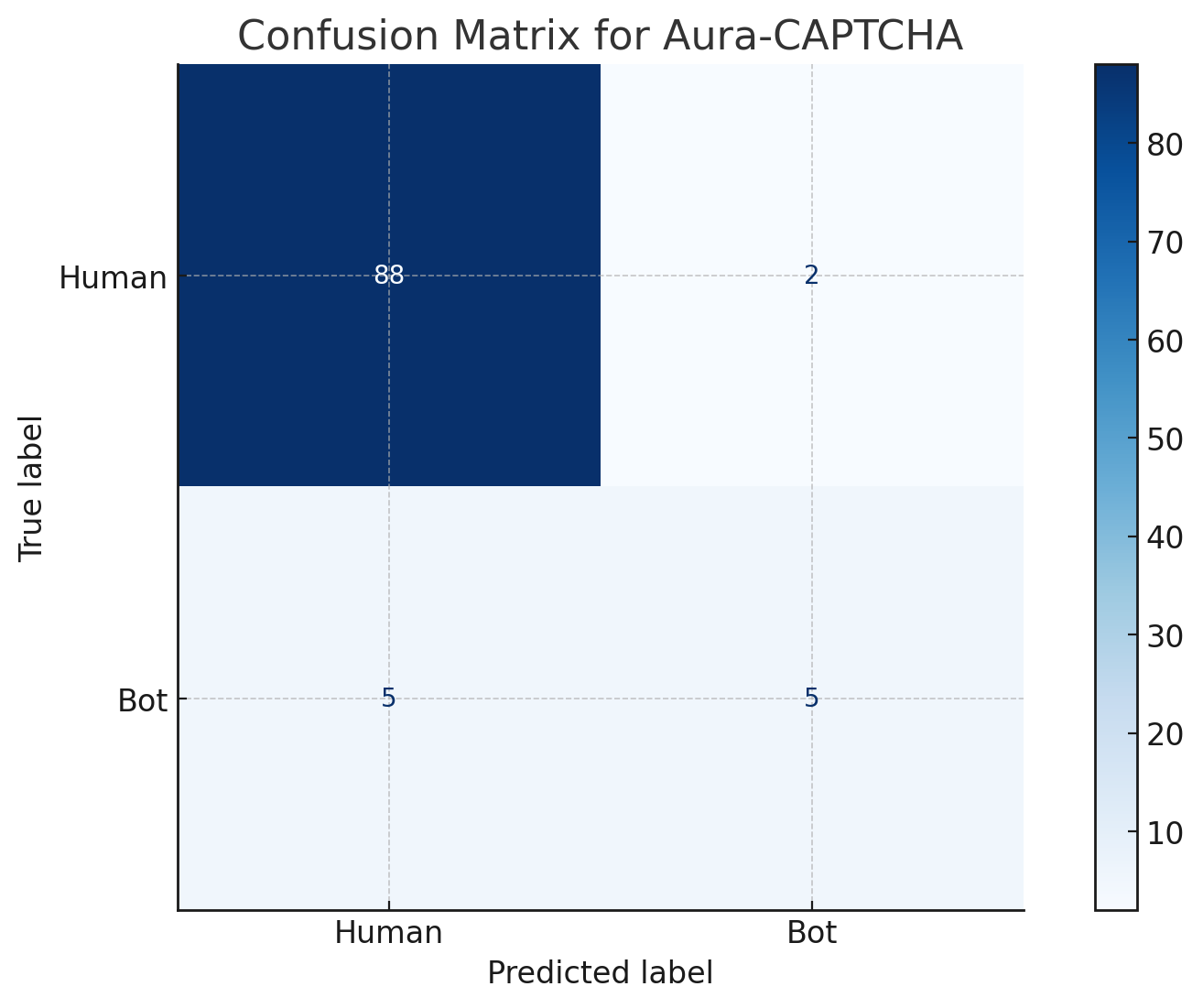}
    \caption{Comparison of CAPTCHA systems across key security and usability metrics. Higher HSR and lower FPR/ART are desirable.}
    \label{fig:comparison}
\end{figure}

Figure~\ref{fig:comparison} visualizes the security-usability trade-off across systems. \textbf{Human Success and Usability.} Aura-CAPTCHA achieves a 92.8\% human success rate with an average response time of 5.6~seconds. These figures compare favorably to static text baselines (78 to 85\% HSR, 8 to 15~s ART) and audio digit baselines (72 to 80\% HSR, 10 to 18~s ART), suggesting that synchronized audio-visual delivery and adaptive difficulty improve the user experience for legitimate humans.

\textbf{Resistance to Classical Attacks.} Against deep-CNN visual solvers, Aura-CAPTCHA's dynamic GAN-based generation yields an 18.4\% bypass rate, substantially lower than the 70.78\% reported by Sivakorn \textit{et al.}~\cite{Sivakorn2016} against static image grids. This improvement stems from the absence of a fixed training distribution: attackers cannot amortize the cost of dataset collection.

Against YOLO-based object detectors, the bypass rate rises to 31.2\%. This is expected because modern object detectors generalize well to novel visual styles; nevertheless, the rate remains below the 100\% achieved by Plesner \textit{et al.}~\cite{Plesner2024} against reCAPTCHA v2 static semantic grids. The combination of abstract geometric patterns (rather than natural-object photographs) and per-session uniqueness contributes to this resistance.

\textbf{Vulnerability to Modern VLM Agents.} Against agentic vision-language models (GPT-4o with browser automation), the bypass rate increases to 58.6\%. This result aligns with the findings of Teoh \textit{et al.}~\cite{Teoh2025}, who demonstrated that VLMs generalize zero-shot across diverse visual CAPTCHA types. We emphasize this limitation transparently: \textit{no explicit visual challenge system, including Aura-CAPTCHA, can claim robustness against state-of-the-art VLMs without additional behavioral or device-integrity signals.} Aura-CAPTCHA mitigates the threat relative to single-modal systems by requiring simultaneous audio reasoning, but multi-modal VLMs are advancing rapidly.

\textbf{Audio Stream Resistance.} Against Whisper ``tiny'' transcribing only the audio channel, the bypass rate is 42.3\%. While this exceeds our visual-only resistance, it reflects the well-documented power of modern ASR~\cite{Radford2023whisper,Hal2025}. Importantly, an attacker must solve \textit{both} modalities to bypass the full system; the joint probability of success is lower than either unimodal rate alone, assuming independence.

\textbf{Interaction Classification.} The hybrid SVM heuristic classifier achieves a 3.1\% false-positive rate, outperforming heuristic-only baselines reported in the literature. Low FPR is critical for accessibility: legitimate users with motor impairments or assistive technologies often exhibit non-standard interaction patterns.

\subsection{Limitations and Threat Model}
\label{subsec:limitations}

We acknowledge the following limitations:
\begin{enumerate}
    \item \textbf{Emerging VLM agents:} As noted, agentic VLMs pose a fundamental challenge to all explicit visual CAPTCHAs. Aura-CAPTCHA reduces but does not eliminate this risk.
    \item \textbf{Scalability of GAN inference:} On-demand GAN generation introduces latency and compute cost compared to serving static images. Our prototype achieves $<$200~ms latency per challenge on a single GPU, but large-scale deployment would require model optimization (e.g., distillation, TensorRT).
    \item \textbf{Behavioral mimicry:} Advanced bots that emulate human-like mouse trajectories and timing can evade heuristic detection. The SVM module improves robustness but is not foolproof against adversaries with access to large-scale human interaction datasets.
    \item \textbf{Comparison with invisible systems:} reCAPTCHA v3 and similar behavioral systems offer lower friction and are the commercial standard. Aura-CAPTCHA is best viewed as a complement, for example, for high-risk login flows or regulatory contexts requiring auditable challenges, rather than a wholesale replacement.
\end{enumerate}

\section{Use Case Applications}
\label{sec:usecases}

Aura-CAPTCHA's adaptive and multi-modal design suits deployment across diverse domains where explicit, auditable human verification is required.

\subsection{E-Commerce and Ticketing}
Online shopping and event-ticketing platforms face scalping bots and fake-account creation. Aura-CAPTCHA can be integrated at account creation, checkout, and review submission. Adaptive difficulty ensures low friction for trusted repeat customers while escalating scrutiny for anomalous sessions.

\subsection{Financial Services}
Banks and payment gateways require transparent, accessible verification for regulatory compliance. Aura-CAPTCHA's multi-modal design supports users with visual impairments (via audio cues) while its interpretable SVM classification aids auditability, which is a requirement absent from opaque black-box risk scores.

\subsection{Social Media and Civic Platforms}
Social media platforms combat bot-driven spam, misinformation, and fake-account campaigns. Aura-CAPTCHA can be deployed during registration or high-volume posting intervals. Dynamic generation prevents adversaries from pre-training solvers on a static challenge corpus.

\subsection{Government and Accessibility-Mandated Services}
Public-sector websites in many jurisdictions must comply with WCAG~2.1 accessibility standards. Aura-CAPTCHA's synchronized audio-visual challenges and adaptive difficulty reduction for struggling users align with inclusive-design principles better than visual-only grids.

\section{Conclusion}
\label{sec:conclusion}

We presented Aura-CAPTCHA, an explicit-challenge CAPTCHA system that integrates GAN-based dynamic content generation, RL-driven adaptive difficulty, and hybrid behavioral classification. In controlled experiments, Aura-CAPTCHA improved human success rates and reduced classical deep-learning bypass rates relative to static text, image, and audio baselines. We reported results transparently, including a 58.6\% bypass rate against modern agentic vision-language models, a threat that challenges the long-term viability of all visual challenge systems.

Our work makes three contributions: (1)~a dual-stream GAN architecture for synchronized audio-visual challenge synthesis; (2)~a Q-learning formulation for real-time CAPTCHA difficulty adaptation; and (3)~empirical characterization of Aura-CAPTCHA against documented SOTA attacks, including honest disclosure of VLM vulnerabilities.

\textbf{Future work.} We identify two urgent directions. First, integrating cognitive-gap tasks that exploit persistent deficits in VLM visual-spatial grounding and state maintenance may restore bot-hardness against agentic solvers. Second, combining dynamic challenges with lightweight device-integrity attestation (e.g., WebAuthn signals) could raise attack costs without sacrificing accessibility. Aura-CAPTCHA provides a stepping stone toward these next-generation defenses.

\bibliographystyle{splncs04}
\bibliography{references}

@inproceedings{vonAhn2003,
  author    = {von Ahn, Luis and Blum, Manuel and Hopper, Nicholas J. and Langford, John},
  title     = {{CAPTCHA}: Using Hard {AI} Problems for Security},
  year      = {2003},
  booktitle = {Advances in Cryptology -- {EUROCRYPT} 2003},
  volume    = {2656},
  series    = {LNCS},
  pages     = {294--311},
  publisher = {Springer}
}

@article{vonAhn2004,
author = {von Ahn, Luis and Blum, Manuel and Langford, John},
title = {Telling humans and computers apart automatically},
year = {2004},
issue_date = {February 2004},
publisher = {Association for Computing Machinery},
address = {New York, NY, USA},
volume = {47},
number = {2},
issn = {0001-0782},
url = {https://doi.org/10.1145/966389.966390},
doi = {10.1145/966389.966390},
journal = {Commun. ACM},
month = feb,
pages = {56–60},
numpages = {5}
}

@inproceedings{Yan2008,
author = {Yan, Jeff and El Ahmad, Ahmad Salah},
title = {A low-cost attack on a Microsoft captcha},
year = {2008},
isbn = {9781595938107},
publisher = {Association for Computing Machinery},
address = {New York, NY, USA},
url = {https://doi.org/10.1145/1455770.1455839},
doi = {10.1145/1455770.1455839},
booktitle = {Proceedings of the 15th ACM Conference on Computer and Communications Security},
pages = {543–554},
numpages = {12},
location = {Alexandria, Virginia, USA},
series = {CCS '08}
}

@article{Bursztein2010sp,
  title={How Good Are Humans at Solving CAPTCHAs? A Large Scale Evaluation},
  author={Elie Bursztein and Steven Bethard and Celine Fabry and John C. Mitchell and Dan Jurafsky},
  journal={2010 IEEE Symposium on Security and Privacy},
  year={2010},
  pages={399-413},
  url={https://api.semanticscholar.org/CorpusID:14204454}
}

@inproceedings{Bursztein2011,
author = {Bursztein, Elie and Martin, Matthieu and Mitchell, John},
title = {Text-based CAPTCHA strengths and weaknesses},
year = {2011},
isbn = {9781450309486},
publisher = {Association for Computing Machinery},
address = {New York, NY, USA},
url = {https://doi.org/10.1145/2046707.2046724},
doi = {10.1145/2046707.2046724},
booktitle = {Proceedings of the 18th ACM Conference on Computer and Communications Security},
pages = {125–138},
numpages = {14},
location = {Chicago, Illinois, USA},
series = {CCS '11}
}

@inproceedings{Fidas2011chi,
author = {Fidas, Christos A. and Voyiatzis, Artemios G. and Avouris, Nikolaos M.},
title = {On the necessity of user-friendly CAPTCHA},
year = {2011},
isbn = {9781450302289},
publisher = {Association for Computing Machinery},
address = {New York, NY, USA},
url = {https://doi.org/10.1145/1978942.1979325},
doi = {10.1145/1978942.1979325},
booktitle = {Proceedings of the SIGCHI Conference on Human Factors in Computing Systems},
pages = {2623–2626},
numpages = {4},
location = {Vancouver, BC, Canada},
series = {CHI '11}
}

@article{Yan2016survey,
author = {Roshanbin, Narges and Miller, James},
title = {A survey and analysis of current CAPTCHA approaches},
year = {2013},
issue_date = {February 2013},
publisher = {Rinton Press, Incorporated},
address = {Paramus, NJ},
volume = {12},
number = {1–2},
issn = {1540-9589},
journal = {J. Web Eng.},
month = feb,
pages = {1–40},
numpages = {40}
}

@misc{Tariq2023,
      title={CAPTCHA Types and Breaking Techniques: Design Issues, Challenges, and Future Research Directions}, 
      author={N. Tariq and F. A. Khan and S. A. Moqurrab and G. Srivastava},
      year={2023},
      eprint={2307.10239},
      archivePrefix={arXiv},
      primaryClass={cs.CR},
      url={https://arxiv.org/abs/2307.10239}, 
}

@misc{Reddy2024,
      title={User Perception of CAPTCHAs: A Comparative Study between University and Internet Users}, 
      author={Arun Reddy and Yuan Cheng},
      year={2024},
      eprint={2405.18547},
      archivePrefix={arXiv},
      primaryClass={cs.CR},
      url={https://arxiv.org/abs/2405.18547}, 
}

@inproceedings{Bursztein2014,
author = {Bursztein, Elie and Aigrain, Jonathan and Moscicki, Angelika and Mitchell, John C.},
title = {The end is nigh: generic solving of text-based CAPTCHAs},
year = {2014},
publisher = {USENIX Association},
address = {USA},
booktitle = {Proceedings of the 8th USENIX Conference on Offensive Technologies},
pages = {3},
numpages = {1},
location = {San Diego, CA},
series = {WOOT'14}
}

@inproceedings{Gao2016ndss,
  title={A Simple Generic Attack on Text Captchas},
  author={Haichang Gao and Jeff Yan and Fang Cao and Zhengya Zhang and Lei Lei and Mengyun Tang and Ping Zhang and Xin Zhou and Xuqin Wang and Jiawei Li},
  booktitle={Network and Distributed System Security Symposium},
  year={2016},
  url={https://api.semanticscholar.org/CorpusID:12024381}
}

@inproceedings{Chellapilla2004nips,
author = {Chellapilla, Kumar and Simard, Patrice Y.},
title = {Using machine learning to break visual human interaction proofs (HIPs)},
year = {2004},
publisher = {MIT Press},
address = {Cambridge, MA, USA},
booktitle = {Proceedings of the 18th International Conference on Neural Information Processing Systems},
pages = {265–272},
numpages = {8},
location = {Vancouver, British Columbia, Canada},
series = {NIPS'04}
}

@article{George2017science,
author = {Dileep George  and Wolfgang Lehrach  and Ken Kansky  and Miguel Lázaro-Gredilla  and Christopher Laan  and Bhaskara Marthi  and Xinghua Lou  and Zhaoshi Meng  and Yi Liu  and Huayan Wang  and Alex Lavin  and D. Scott Phoenix },
title = {A generative vision model that trains with high data efficiency and breaks text-based CAPTCHAs},
journal = {Science},
volume = {358},
number = {6368},
pages = {eaag2612},
year = {2017},
doi = {10.1126/science.aag2612},
URL = {https://www.science.org/doi/abs/10.1126/science.aag2612},
eprint = {https://www.science.org/doi/pdf/10.1126/science.aag2612},
}

@article{Sivakorn2016,
  title={I am Robot: (Deep) Learning to Break Semantic Image CAPTCHAs},
  author={Suphannee Sivakorn and Iasonas Polakis and Angelos Dennis Keromytis},
  journal={2016 IEEE European Symposium on Security and Privacy (EuroS\&P)},
  year={2016},
  pages={388-403},
  url={https://api.semanticscholar.org/CorpusID:206649515}
}

@misc{Hossen2020,
      title={An Object Detection based Solver for Google's Image reCAPTCHA v2}, 
      author={Md Imran Hossen and Yazhou Tu and Md Fazle Rabby and Md Nazmul Islam and Hui Cao and Xiali Hei},
      year={2021},
      eprint={2104.03366},
      archivePrefix={arXiv},
      primaryClass={cs.CR},
      url={https://arxiv.org/abs/2104.03366}, 
}

@inproceedings{Plesner2024,
   title={Breaking reCAPTCHAv2},
   url={http://dx.doi.org/10.1109/COMPSAC61105.2024.00142},
   DOI={10.1109/compsac61105.2024.00142},
   booktitle={2024 IEEE 48th Annual Computers, Software, and Applications Conference (COMPSAC)},
   publisher={IEEE},
   author={Plesner, Andreas and Vontobel, Tobias and Wattenhofer, Roger},
   year={2024},
   pages={1047–1056} 
   }

@inproceedings{Bursztein2009,
author = {Bursztein, Elie and Bethard, Steven},
title = {Decaptcha: breaking 75\% of eBay audio CAPTCHAs},
year = {2009},
publisher = {USENIX Association},
address = {USA},
booktitle = {Proceedings of the 3rd USENIX Conference on Offensive Technologies},
pages = {8},
numpages = {1},
location = {Montreal, Canada},
series = {WOOT'09}
}

@inproceedings{Bursztein2011audio,
title = {The failure of noise-based non-continuous audio captchas},
author = {"Elie, Bursztein" and "Romain, Bauxis" and "Hristo, Paskov" and "Daniele, Perito" and "Celine, Fabry" and "John C., Mitchell"},
booktitle = {Security and Privacy},
year = {2011},
organization = {IEEE}
}

@inproceedings{Bock2017,
author = {Bock, Kevin and Patel, Daven and Hughey, George and Levin, Dave},
title = {unCaptcha: a low-resource defeat of recaptcha's audio challenge},
year = {2017},
publisher = {USENIX Association},
address = {USA},
booktitle = {Proceedings of the 11th USENIX Conference on Offensive Technologies},
pages = {7},
numpages = {1},
location = {Vancouver, BC, Canada},
series = {WOOT'17}
}

@inproceedings{Solanki2017,
author = {Solanki, Saumya and Krishnan, Gautam and Sampath, Varshini and Polakis, Jason},
title = {In (Cyber)Space Bots Can Hear You Speak: Breaking Audio CAPTCHAs Using OTS Speech Recognition},
year = {2017},
isbn = {9781450352024},
publisher = {Association for Computing Machinery},
address = {New York, NY, USA},
url = {https://doi.org/10.1145/3128572.3140443},
doi = {10.1145/3128572.3140443},
booktitle = {Proceedings of the 10th ACM Workshop on Artificial Intelligence and Security},
pages = {69–80},
numpages = {12},
location = {Dallas, Texas, USA},
series = {AISec '17}
}

@article{Sano2015,
  title={HMM-based Attacks on Google's ReCAPTCHA with Continuous Visual and Audio Symbols},
  author={Shotaro Sano and Takuma Otsuka and Katsutoshi Itoyama and Hiroshi G. Okuno},
  journal={J. Inf. Process.},
  year={2015},
  volume={23},
  pages={814-826},
  url={https://api.semanticscholar.org/CorpusID:46183584}
}

@misc{Radford2023whisper,
      title={Robust Speech Recognition via Large-Scale Weak Supervision}, 
      author={Alec Radford and Jong Wook Kim and Tao Xu and Greg Brockman and Christine McLeavey and Ilya Sutskever},
      year={2022},
      eprint={2212.04356},
      archivePrefix={arXiv},
      primaryClass={eess.AS},
      url={https://arxiv.org/abs/2212.04356}, 
}

@inproceedings{Hal2025,
  TITLE = {{Bypassing Audio reCAPTCHA with Automatic Speech Recognition Models}},
  AUTHOR = {Aubry, Paul and Devoivre, Juliette and Carron, Damien and Fernandez, Simon and Duda, Andrzej and Korczy{\'n}ski, Maciej},
  URL = {https://hal.science/hal-05489792},
  BOOKTITLE = {{2025 IEEE European Symposium on Security and Privacy Workshops (EuroS\&amp;PW)}},
  ADDRESS = {Venice, Italy},
  PUBLISHER = {{IEEE}},
  PAGES = {1-5},
  YEAR = {2025},
  MONTH = Jun,
  DOI = {10.1109/EuroSPW67616.2025.00032},
  HAL_ID = {hal-05489792},
  HAL_VERSION = {v1},
}

@inproceedings{Teoh2025,
author = {Teoh, Xiwen and Lin, Yun and Li, Siqi and Liu, Ruofan and Sollomoni, Avi and Harel, Yaniv and Dong, Jin Song},
title = {Are CAPTCHAs still bot-hard? generalized visual CAPTCHA solving with agentic vision language model},
year = {2025},
isbn = {978-1-939133-52-6},
publisher = {USENIX Association},
address = {USA},
booktitle = {Proceedings of the 34th USENIX Conference on Security Symposium},
articleno = {193},
numpages = {20},
location = {Seattle, WA, USA},
series = {SEC '25}
}

@inproceedings{Teoh2024,
author = {Teoh, Xiwen and Lin, Yun and Liu, Ruofan and Huang, Zhiyong and Dong, Jin Song},
title = {PhishDecloaker: detecting CAPTCHA-cloaked phishing websites via hybrid vision-based interactive models},
year = {2024},
isbn = {978-1-939133-44-1},
publisher = {USENIX Association},
address = {USA},
booktitle = {Proceedings of the 33rd USENIX Conference on Security Symposium},
articleno = {29},
numpages = {18},
location = {Philadelphia, PA, USA},
series = {SEC '24}
}

@misc{Shi2019,
      title={Adversarial CAPTCHAs}, 
      author={Chenghui Shi and Xiaogang Xu and Shouling Ji and Kai Bu and Jianhai Chen and Raheem Beyah and Ting Wang},
      year={2019},
      eprint={1901.01107},
      archivePrefix={arXiv},
      primaryClass={cs.CR},
      url={https://arxiv.org/abs/1901.01107}, 
}

@article{Wang2023audioadv,
author={Wang, Ping and Gao, Haichang and Guo, Xiaoyan and Yuan, Zhongni and Nian, Jiawei},
journal={ IEEE Transactions on Dependable and Secure Computing },
title={{ Improving the Security of Audio CAPTCHAs With Adversarial Examples }},
year={2024},
volume={21},
number={02},
ISSN={1941-0018},
pages={650-667},
doi={10.1109/TDSC.2023.3236367},
url = {https://doi.ieeecomputersociety.org/10.1109/TDSC.2023.3236367},
publisher={IEEE Computer Society},
address={Los Alamitos, CA, USA}
}

@inproceedings{Gao2013kdd,
author = {Moradi, Morteza and Moradi, Mohammad and Palazzo, Simone and Borji, Ali and Spampinato, Concetto},
title = {Camouflaged in Patterns: Designing Robust and User-Friendly CAPTCHA Challenges via Texture Blending},
year = {2025},
isbn = {9798400721021},
publisher = {Association for Computing Machinery},
address = {New York, NY, USA},
url = {https://doi.org/10.1145/3750069.3755953},
doi = {10.1145/3750069.3755953},
booktitle = {Proceedings of the 16th Biannual Conference of the Italian SIGCHI Chapter},
articleno = {95},
numpages = {3},
location = {
},
series = {CHItaly '25}
}

@inproceedings{Ross2010,
author = {Ross, Steven A. and Halderman, J. Alex and Finkelstein, Adam},
title = {Sketcha: a captcha based on line drawings of 3D models},
year = {2010},
isbn = {9781605587998},
publisher = {Association for Computing Machinery},
address = {New York, NY, USA},
url = {https://doi.org/10.1145/1772690.1772774},
doi = {10.1145/1772690.1772774},
booktitle = {Proceedings of the 19th International Conference on World Wide Web},
pages = {821–830},
numpages = {10},
location = {Raleigh, North Carolina, USA},
series = {WWW '10}
}

@inproceedings{Elson2007,
author = {Elson, Jeremy and Douceur, John (JD) and Howell, Jon and Saul, Jared},
title = {Asirra: A CAPTCHA that Exploits Interest-Aligned Manual Image Categorization},
booktitle = {Proceedings of 14th ACM Conference on Computer and Communications Security (CCS)},
year = {2007},
month = {October},
publisher = {Association for Computing Machinery, Inc.},
url = {https://www.microsoft.com/en-us/research/publication/asirra-a-captcha-that-exploits-interest-aligned-manual-image-categorization/},
edition = {Proceedings of 14th ACM Conference on Computer and Communications Security (CCS)},
}

@inproceedings{Wang2018icassp,
author = {Wang, Haipeng and Zheng, Feng and Chen, Zhuoming and Lu, Yi and Gao, Jing and Wei, Renjia},
title = {A Captcha Design Based on Visual Reasoning},
year = {2018},
publisher = {IEEE Press},
url = {https://doi.org/10.1109/ICASSP.2018.8461764},
doi = {10.1109/ICASSP.2018.8461764},
booktitle = {2018 IEEE International Conference on Acoustics, Speech and Signal Processing (ICASSP)},
pages = {1967–1971},
numpages = {5},
location = {Calgary, AB, Canada}
}

@article{Lee2023mtap,
  author = {Lee, J.H., et al.},
  title = {Stabilized Performance Maximization for GAN-based Real-Time Authentication Image Generation over Internet.},
  year = {2023},
  journal = {Multimedia Tools and Applications},
  volume = {82},
  pages = {15885-15908}
}

@misc{Li2023cyclegan,
       title={An End-to-End Attack on Text-based CAPTCHAs Based on Cycle-Consistent Generative Adversarial Network}, 
      author={Chunhui Li and Xingshu Chen and Haizhou Wang and Yu Zhang and Peiming Wang},
      year={2020},
      eprint={2008.11603},
      archivePrefix={arXiv},
      primaryClass={cs.CV},
      url={https://arxiv.org/abs/2008.11603}, 
}

@misc{Shet2014,
  author = {Shet, V.},
  title = {Are You a Robot? Introducing ``No CAPTCHA reCAPTCHA''.},
  year = {2014},
  howpublished = {Google Online Security Blog (2014)}
}

@misc{Prince2020,
  author = {Prince, Matthew and Isasi, Sergi},
  title = {Moving from reCAPTCHA to hCaptcha},
  year = {2020},
  howpublished = {Cloudflare Blog},
  url = {https://blog.cloudflare.com/moving-from-recaptcha-to-hcaptcha/}
}

@misc{imperva2024report,
  author = {{Imperva}},
  title = {Bad Bot Report 2024},
  year = {2024},
  howpublished = {\url{https://www.imperva.com/resources/resource-library/reports/2024-imperva-bad-bot-report/}}
}

\end{document}